\documentclass[runningheads]{llncs}
\usepackage{graphicx}
\usepackage{comment}
\usepackage{epsfig}
\usepackage{amsmath,amssymb} % define this before the line numbering.
\usepackage{color}
\usepackage[table]{xcolor}
\usepackage{tikz}
\usepackage{comment}
\usepackage{multirow}
\usepackage[hang,flushmargin]{footmisc} 
\usepackage{booktabs,tabularx}
\usepackage{enumitem}
\usepackage{makecell}
\usepackage{algorithm}
\usepackage{algorithmic}
\usepackage{authblk}
\usepackage{threeparttable}
\usepackage{lipsum}
\usepackage[pagebackref=false, breaklinks=true, colorlinks, bookmarks=false]{hyperref}
\usepackage{url}

\usepackage[width=122mm,left=12mm,paperwidth=146mm,height=193mm,top=12mm,paperheight=217mm]{geometry}
\usepackage{cite}
\newcommand{\fig}[1]{Fig. \ref{#1}}

\newcommand{\tab}[1]{Table \ref{#1}}

\newcommand\blfootnote[1]{%
\begingroup
\renewcommand\thefootnote{}\footnote{#1}%
\addtocounter{footnote}{-1}%
\endgroup
}

\begin{document}
% \linenumbers
\pagestyle{headings}
\mainmatter
\def\ECCVSubNumber{685}  % Insert your submission number here

\title{When Counting Meets HMER:\\ Counting-Aware Network for Handwritten Mathematical Expression Recognition} % Replace with your title

%\begin{comment}
% \titlerunning{ECCV-22 submission ID \ECCVSubNumber} 
% \authorrunning{ECCV-22 submission ID \ECCVSubNumber} 
% \author{Anonymous ECCV submission}
% \institute{Paper ID \ECCVSubNumber}
%\end{comment}
%******************

% \begin{comment}
\titlerunning{When Counting Meets HMER: Counting-Aware Network for HMER}
\author{Bohan Li\inst{1,2}$^{\ast}$ \and Ye Yuan\inst{1}$^{\ast}$ \and Dingkang Liang\inst{2} \and Xiao Liu\inst{1} \and Zhilong Ji\inst{1} \and \linebreak Jinfeng Bai\inst{1} \and Wenyu Liu\inst{2} \and Xiang Bai\inst{2}$^{\ddagger}$}
\authorrunning{B. Li et al.}
\institute{Tomorrow Advancing Life \\
\email{\{yuanye\_phy\}@hotmail.com,\{jizhilong\}@tal.com\\
\{ender.liux,jfbai.bit\}@gmail.com} \and Huazhong University of Science and Technology \email{\{bohan1024,dkliang,liuwy,xbai\}@hust.edu.cn}}
% \end{comment}
%******************
\maketitle

\begin{abstract}

Recently, most handwritten mathematical expression recognition (HMER) methods adopt the encoder-decoder networks, which directly predict the markup sequences from formula images with the attention mechanism. However, such methods may fail to accurately read formulas with complicated structure or generate long markup sequences, as the attention results are often inaccurate due to the large variance of writing styles or spatial layouts. To alleviate this problem, we propose an unconventional network for HMER named Counting-Aware Network (CAN), which jointly optimizes two tasks: HMER and symbol counting. Specifically, we design a weakly-supervised counting module that can predict the number of each symbol class without the symbol-level position annotations, and then plug it into a typical attention-based encoder-decoder model for HMER. Experiments on the benchmark datasets for HMER validate that both joint optimization and counting results are beneficial for correcting the prediction errors of encoder-decoder models, and CAN consistently outperforms the state-of-the-art methods. In particular, compared with an encoder-decoder model for HMER, the extra time cost caused by the proposed counting module is marginal. The source code is available at \url{https://github.com/LBH1024/CAN}.\blfootnote{$^{\ast}$Authors contribute equally. $^{\ddagger}$Corresponding author.}

%Existing Handwritten Mathematical Expression Recognition (HMER) methods mostly use the attention mechanism to automatically learn where to focus at every step. Though effective to a certain extent, the existing attention-based frameworks still struggle to handle the complexity of HMER, where the symbols are in complicated spatial layout and the relation between two adjacent symbols is position-sensitive. To address this issue, in this paper we creatively bring multi-class symbol counting into HMER, which can make the model develop a global awareness of each symbol's position. This global awareness can implicitly serve as a guided map for the one-symbol-to-another attention process, which is troubled by the complicated spatial layout. Specifically, we designs a counting module to predict the count of each class of symbols, and then plug it into an attention-based encoder-decoder framework to propose an unified end-to-end trainable framework named \textbf{C}ounting-\textbf{A}ware \textbf{N}etwork (CAN). In this framework we also utilize the counting result as a global complement to the local feature to further improve the performance. Extensive experiments on several public datasets demonstrate the superiority of CAN over the state-of-the-art methods. Beyond this, we also discuss the mutual impact of multi-class symbol counting and HMER, which paves the way for future related research.
\keywords{Handwritten mathematical expression recognition \and Attention mechanism \and Counting}
\end{abstract}

\section{Introduction}

Handwritten mathematical expression recognition (HMER) is an important task of document analysis, which has broad applications including assignment grading, digital library service, and office automation. Despite the great successes of the current OCR systems, HMER still remains a very challenging problem due to the complex structures of formulas or irregular writings.  

Encoder-decoder architectures are extensively used in the recent HMER approaches \cite{DBLP:conf/icpr/ZhangDD18,DBLP:journals/ijcv/WuYZZL20,abm}, which formulate HMER as an image-to-sequence translation problem. Given a handwritten formula, such methods predict its corresponding markup sequence (e.g., LaTeX) with the attention mechanism. However, encoder-decoder models often cannot guarantee the accuracy of attention, especially when the structure of a handwritten formula is complicated or the markup sequence is long.  

\begin{figure*}[t]
\begin{center}
  \includegraphics[width=0.8\linewidth]{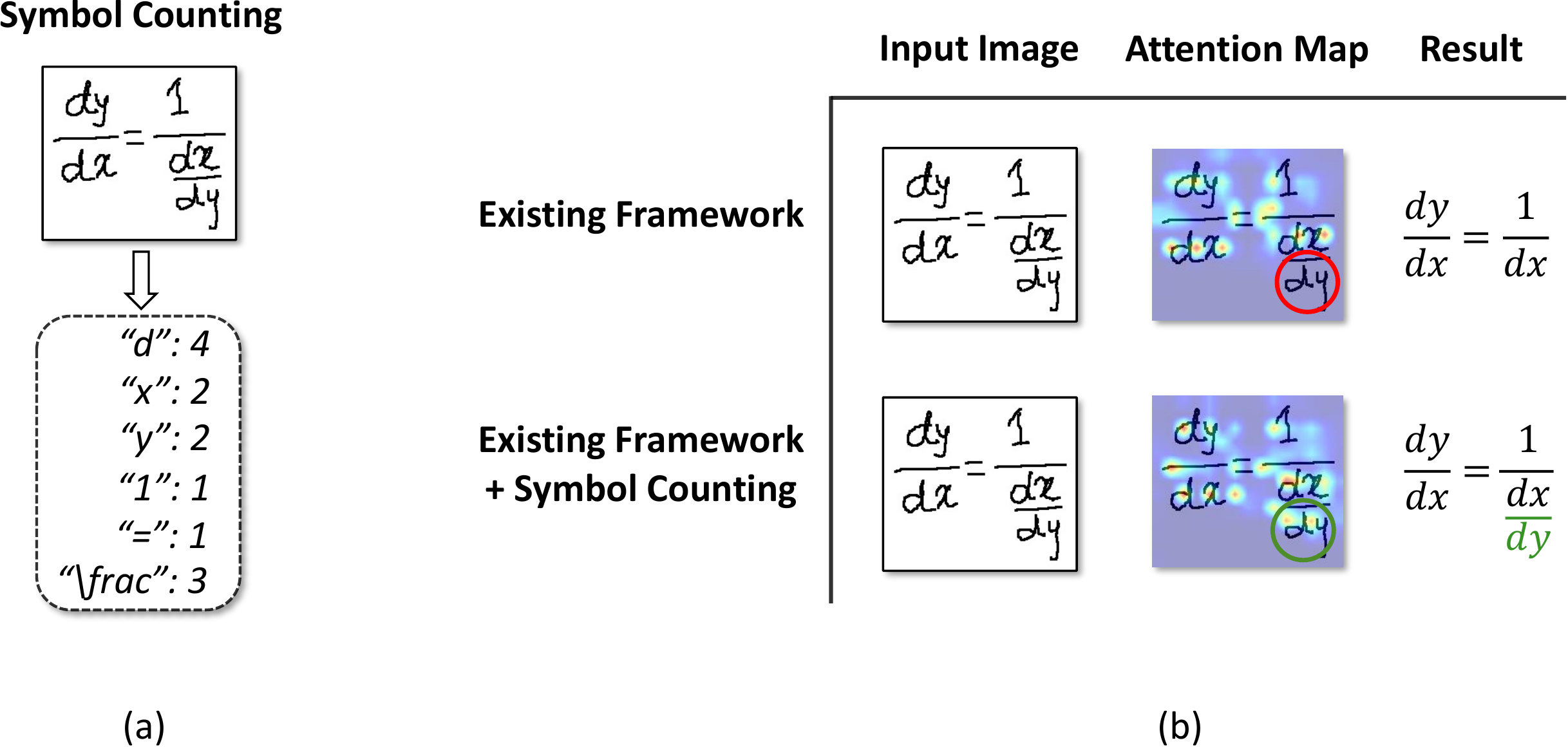}
\end{center}
\vspace{-1ex}
\caption{
%(a) is the illustration of the symbol counting task. (b) is the comparison between the existing framework and our CAN.
(a) Illustration of the symbol counting task. (b) Comparison between the existing framework (e.g., DWAP~\cite{DBLP:conf/icpr/ZhangDD18}) and the proposed framework (CAN). By visualizing the attention map, we can observe that the existing framework misses the denominator $``dy”$ while our CAN correctly locate it after using symbol counting.}
\vspace{-1ex}
\label{intro}
\end{figure*}

% 在这篇文章中，我们提出了一种非常规的方法，用来提升encoder-decoder的鲁棒性。我们认为，counting和HMER具有互补性，并且可以使用counting来提升HMER的性能 。In this community，object counting （引文献）has been intensively studied, but has seldom been applied in OCR area. 我们的intuition是1）symbol counting（as illustrated in fig1）能给encoder和decoder提供字符的位置信息，这种位置信息能够增强attention的准确性 2）counting的结果（公式里面的字符个数）可以作为一种全局的额外的信息（representation）来提升识别的准确性。

In this paper, we propose an unconventional method for improving the robustness of the encoder-decoder models for HMER. We argue that counting and HMER are two complementary tasks, and using counting can improve the performance of HMER. In this community, object counting~\cite{li2018csrnet,laradji2018blobs} has been intensively studied, but has seldom been applied in the OCR area. Our intuition includes two aspects: 1) symbol counting (as illustrated in \fig{intro}(a)) is able to provide the symbol-level position information, which can make the attention results more accurate. 2) The counting results, representing the number of each symbol class, can serve as additional global information to promote recognition accuracy.    

% 具体来说，我们设计了一个weakly-supervised symbol counting module, which is able to plug into existing encoder-decoder networks 并且能端到端一起联合训练。With this counting module, encoder-decoder 模型能够更加准确感受到字符，如fig1b所示。需要强调的是，所提出的counting module不需要symbol-level position annotations，只需要use the original HEMR annotations。我们在DWAP（以它为baseline）结合我们的counting module提出了一个新的公式识别的方法，在标准数据集上面我们测试了我们的方法，可以观察到both HMER and symbol counting 都可以获得obvious improvement。特别地，计数模块带来的时间开销相比较于原来的encoder-decoder所需要的的时间开销，可以忽略不计。

Specifically, we design a weakly-supervised counting module named MSCM, which can be easily plugged into existing encoder-decoder networks and optimized jointly in an end-to-end manner. With this counting module, an encoder-decoder model can be better aware of each symbol's position, as shown in \fig{intro}(b). It is worth noticing that the proposed counting module just needs original HMER annotations (LaTeX sequences) without extra labeling work. We combine our counting module with a typical encoder-decoder model (e.g., DWAP~\cite{DBLP:conf/icpr/ZhangDD18}), proposing a unified network for HMER named Counting-Aware Network (CAN). We test it on the benchmark datasets and observe that both HMER and symbol counting gain obvious and consistent performance improvement. In particular, compared with the original model, the extra time cost brought by MSCM is marginal.  

% 总结，贡献：
% 1.	我们首次将计数引入到HMER中并且reveal了手写公式识别和计数之间的关联和互补性。 
% 2.	我们设计了一个新的方法，它可以将计数任务和HMER进行联合优化并且端到端训练，取得了计数任务和HMER性能上一致稳定的提升。

In summary, the main contributions of this paper are two-fold. 1) To the best of our knowledge, we are the first to bring symbol counting into HMER and reveal the relevance and the complementarity of HMER and symbol counting. 2) We propose a new method that jointly optimizes symbol counting and HMER, which consistently improves the performance of the encoder-decoder models for HMER.

% 以wap为baseline，our method 在 xxx 数据集上面取得了一致的sota的结果。进一步的，我们将这个模块结合了最新的HMER方法，也依然取得了xxx的提升。这证明了我们的方法具有通用性，可以用来改进现有的大部分encoder-decoder方法。文章剩下部分的组织。

To be specific about the performance, with adopting DWAP~\cite{DBLP:conf/icpr/ZhangDD18} as the baseline network, our method achieves state-of-the-art (SOTA) recognition accuracy on the widely-used CROHME dataset (57.00\% on CROHME 2014, 56.06\% on CROHME 2016, 54.88\% on CROHME 2019). Moreover, with adopting the latest SOTA method ABM~\cite{abm} as the baseline network, CAN achieves new SOTA results (57.26\% on CROHME 2014, 56.15\% on CROHME 2016, 55.96\% on CROHME 2019). This indicates that our method can be generalized to various existing encoder-decoder models for HMER and boost their performance.

\section{Related Work}\label{related work}
\subsection{HMER}
Traditional HMER methods usually take a three-step approach: a symbol segmentation step, a symbol recognition step, and a grammar-guided structure analysis step. Classic classification techniques such as HMM \cite{winkler1996hmm,kosmala1999line,hu2011hmm}, Elastic Matching \cite{chan1998elastic,vuong2010towards} and Support Vector Machines \cite{keshari2007hybrid} are mainly used in the recognition step. In the structure analysis step, formal grammars are elaborately designed to model the 2D and syntactic structures of formulas. Lavirotte \emph{et al.} \cite{lavirotte1998mathematical} propose to use graph grammar to recognize mathematical expression. Chan \emph{et al.} \cite{chan2001error} incorporate correction mechanism into a parser based on definite clause grammar (DCG). However, limited feature learning ability and complex grammar rules make the traditional methods far to meet real-world application.

Recently, deep learning has rapidly boosted the performance of HMER. The mainstream framework is the encoder-decoder network\cite{DBLP:conf/icml/ZhangDYSW020,DBLP:conf/icfhr/TruongNPN20,DBLP:conf/pkdd/WuYZZL18,DBLP:journals/ijcv/WuYZZL20,abm,DBLP:conf/icpr/ZhangDD18,DBLP:conf/icml/DengKLR17,DBLP:conf/icdar/ZhaoGYPDZ21,DBLP:journals/pr/ZhangDZLHHWD17}. Deng \emph{et al.} \cite{DBLP:conf/icml/DengKLR17} first apply an attention-based encoder-decoder model in HMER, inspired by its success in image caption task \cite{DBLP:conf/icml/XuBKCCSZB15}. Zhang \emph{et al.} \cite{DBLP:journals/pr/ZhangDZLHHWD17} also present a similar model named WAP. In their model, they apply a FCN as the encoder and utilize the coverage attention, which is the sum of all past attention weights, to alleviate the lack of coverage problem \cite{DBLP:conf/acl/TuLLLL16}. Wu \emph{et al.} \cite{DBLP:conf/pkdd/WuYZZL18,DBLP:journals/ijcv/WuYZZL20} focus on the pair-wise adversarial learning strategy to improve the recognition accuracy. Later, Zhang \emph{et al.} \cite{DBLP:conf/icml/ZhangDYSW020} devise a tree-based decoder to parse formulas. At each step, a parent and child node pair is generated and the relation between parent node and child node reflects the structure type. Bi-directional learning has been proven effective to improve model recognition performance \cite{shi2018aster}. Zhao \emph{et al.} \cite{DBLP:conf/icdar/ZhaoGYPDZ21} design a bi-directionally trained transformer framework and Bian \emph{et al.} \cite{abm} propose an bi-directional mutual learning network. They further prove bi-directional learning can also significantly improve the HMER performance. 

\subsection{Object Counting}
Object counting can be roughly divided into two categories, detection-based and regression-based. The detection-based methods~\cite{ren2015faster,liu2016ssd} obtain the number by detecting each instance. The regression-based methods~\cite{li2018csrnet,xu2022autoscale} learn to count by regressing a density map, and the predicted count equals the integration of the density map. To improve the counting accuracy, multi-scale strategy~\cite{zhang2016single}, attention mechanism~\cite{zhang2019attentional} and perspective information~\cite{yan2019perspective} are widely adopted in the regression-based methods. Nevertheless, both detection-based and density map regression-based methods need the object position annotations (fully-supervised), such as box-level~\cite{ren2015faster,liu2016ssd} and point-level~\cite{zhang2016single,xu2022autoscale,li2018csrnet} annotations. To relieve the expensive and laborious labeling work, several approaches~\cite{yang2020weakly,wang2015deep} that only use count-level annotations (weakly-supervised) are proposed. And they find that the visualized feature map can accurately reflect the object regions.
% Specifically, Yang~\textit{et al.}~\cite{yang2020weakly} proposes a soft-label sorting network to directly regress the object number. Wang~\textit{et al.}~\cite{wang2015deep} also directly maps the image to a count, using some negative samples to boost the robustness. Liang~\textit{et al.}~\cite{liang2021transcrowd} formulates the counting problem as sequence-to-count paradigm. 
Different from most of the previous counting modules that are category specifically (e.g., crowd counting), our counting module is designed for multi-class object counting since formulas usually contains various symbols. In the OCR area, Xie \emph{et al.} \cite{xie2019aggregation} propose a counting-based loss function mainly designed
for scene texts (words or text-lines), while our model can
exploit the counting information of more complicated texts
(e.g., mathematical expressions) at both the feature level
and the loss level.

\begin{figure*}[t]
\begin{center}
  \includegraphics[width=0.90\linewidth]{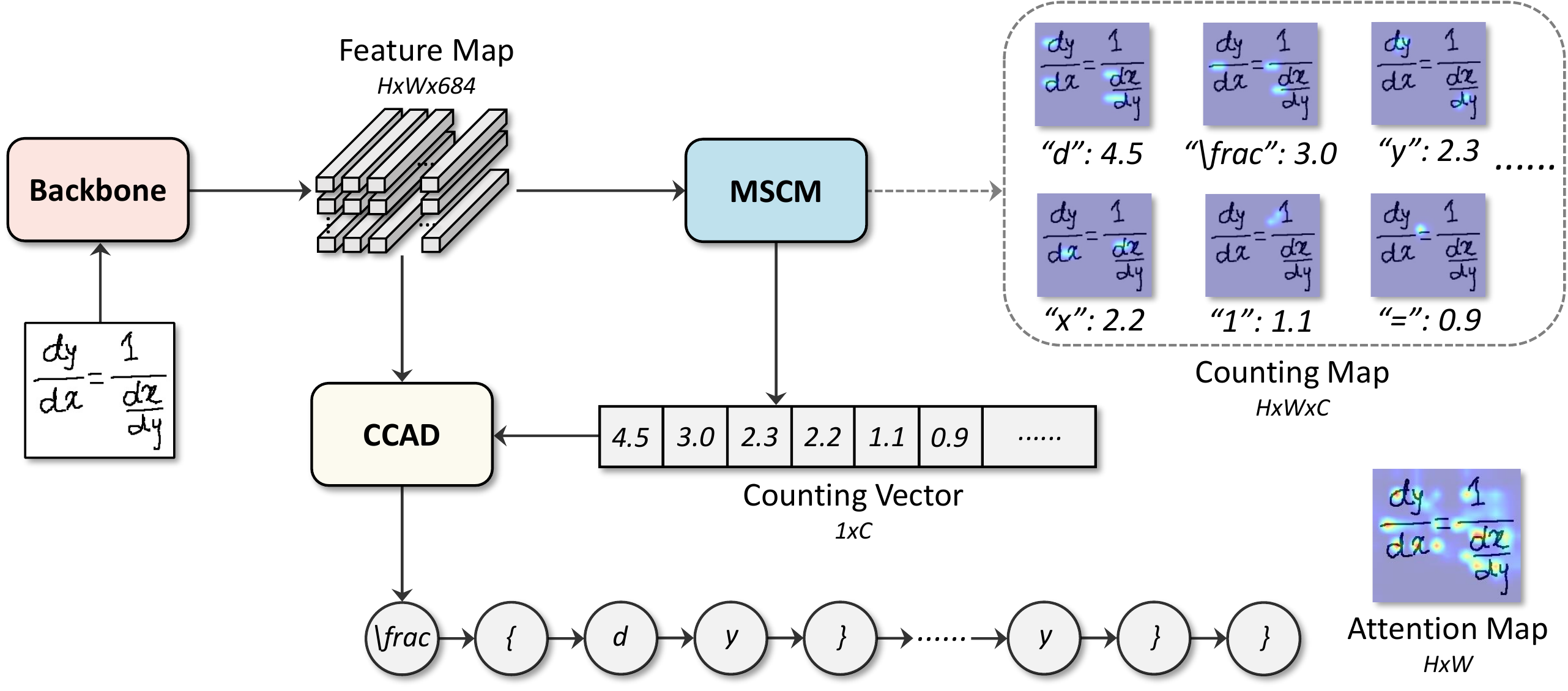}
\end{center}
\vspace{-1ex}
\caption{Structure of the proposed CAN, which consists of a backbone network, a multi-scale counting Module (MSCM) and a counting-combined attentional decoder (CCAD).}
\label{main}
\end{figure*}

\section{Methodology}\label{method}

\subsection{Overview}

As shown in \fig{main}, our Counting-Aware Network (CAN) is a unified end-to-end trainable framework that comprises a backbone, a multi-scale counting module (MSCM) and a counting-combined attentional decoder (CCAD). Following DWAP~\cite{DBLP:conf/icpr/ZhangDD18}, we apply DenseNet~\cite{DBLP:conf/cvpr/HuangLMW17} as the backbone. Given a gray-scale image $\mathcal{X} \in \mathbb{R}^{H^{'} \times W^{'} \times 1}$, the backbone is first used to extract 2D feature map $\mathcal{F} \in \mathbb{R}^{H \times W \times 684}$, where $\frac{H^{'}}{H} = \frac{W^{'}}{W} = 16$.   
The feature map $\mathcal{F}$ will be used by both the MSCM and the CCAD. The counting module MSCM is used to predict the number of each symbol class and generate the 1D counting vector $\mathcal{V}$ that represents the counting results. The feature map $\mathcal{F}$ and the counting vector $\mathcal{V}$ will be fed into the CCAD to get predicted output. 

\subsection{Multi-Scale Counting Module}\label{mscm}
\begin{figure*}[t]
\begin{center}
  \includegraphics[width=0.96\linewidth]{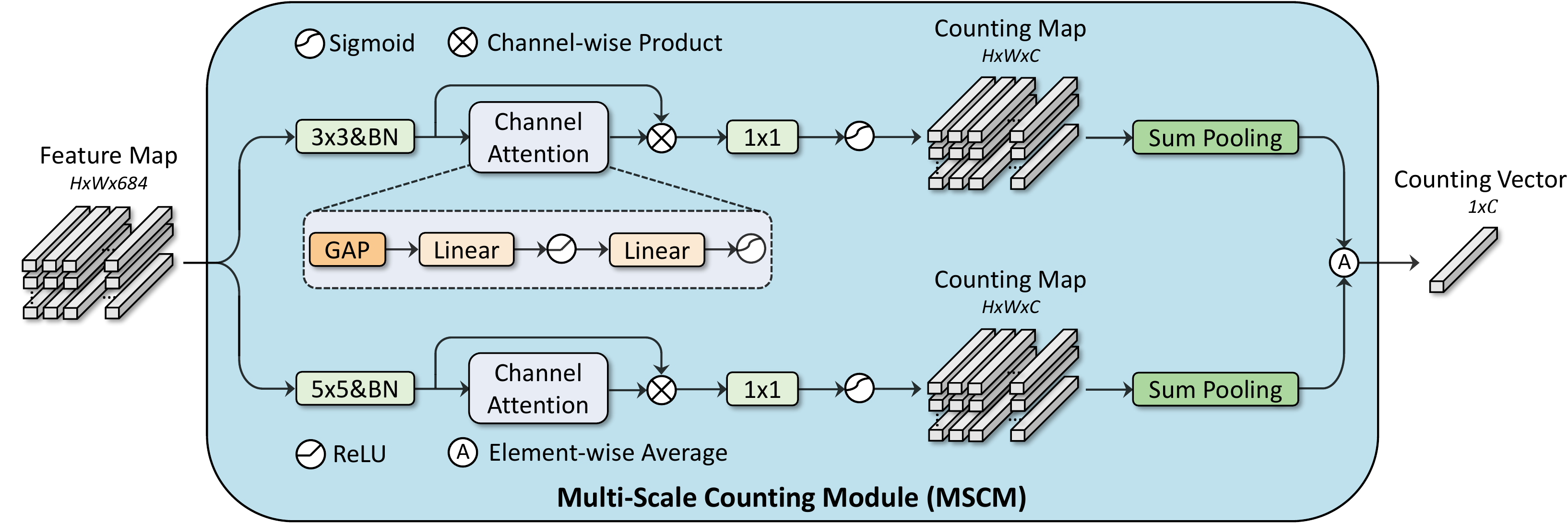}
\end{center}
\vspace{-1ex}
\caption{Structure of the proposed multi-scale counting module (MSCM).}
\label{MSCM}
\end{figure*}

In this part, we present the detail of the proposed multi-scale counting module (MSCM), which is designed to predict the number of each symbol class. Specifically, as depicted in \fig{MSCM}, MSCM consists of multi-scale feature extraction, channel attention and sum-pooling operator. Formula images usually contain various sizes of symbols due to different writing habits. Single kernel size can not effectively handle the scale variations. To this end, we first utilize two parallel convolution branches to extract multi-scale features by using different kernel sizes (set to 3 $\times$ 3 and 5 $\times$ 5). Following the convolution layer, the channel attention~\cite{DBLP:conf/cvpr/HuSS18} is adopted to enhance the feature information further. Here, we choose one of the branches for simple illustration. Let us denote $\mathcal{H} \in \mathbb{R}^{H \times W \times C^{'}}$ as the extracted feature map from the convolution (3 $\times$ 3 or 5 $\times$ 5) layer. The enhanced feature $S$ can be written as:
\begin{equation}
\label{eq1}
\displaystyle{\mathcal{Q}=\sigma(W_{1}(G(\mathcal{H}))+b_{1}), }
\vspace{-2.5ex}
\end{equation}

\begin{equation}
\label{eq2}
\displaystyle{\mathcal{S}=\mathcal{Q} \otimes g(W_{2}\mathcal{Q}+b_{2}), }
\vspace{+1.5ex}
\end{equation}
where $G$ is the global average pooling. $\sigma$ and $g(\cdot)$ refer to ReLU and sigmoid function, respectively. $\otimes$ denotes channel-wise product and $W_{1}$, $W_{2}$, $b_{1}$, $b_{2}$ are trainable weights.

After getting the enhanced feature $\mathcal{S}$, we use a $1 \times 1$ convolution to reduce the channel number from $C^{'}$ to $C$, where C is the number of symbol classes. Ideally, the symbol counting result should mainly calculate from the foreground (symbols), i.e., the response of the background should be close to zero. Thus, following the $1 \times 1$ convolution, we utilize a sigmoid function to yield the value in a range of (0,1) to generate counting map $\mathcal{M} \in \mathbb{R}^{H \times W \times C}$. For each $\mathcal{M}_{i} \in \mathbb{R}^{H \times W}$, it is supposed to effectively reflect the position of the $i$-th symbol class, as shown in \fig{main}. In this sense, each $\mathcal{M}_{i}$ is actually a pseudo density map, and we can utilize sum-pooling operator to obtain counting vector $\mathcal{V} \in \mathbb{R}^{1 \times C}$:
% \vspace{-1ex}
\begin{equation}
\label{eq3}
\displaystyle{\mathcal{V}_{i}=\sum_{p=1}^{H}\sum_{q=1}^{W}M_{i,pq}}
% \displaystyle{\mathcal{V}_{i}=\text{sum}(M_{i}), }
% \vspace{-1ex}
\end{equation}

Here, $\mathcal{V}_{i} \in \mathbb{R}^{1 \times 1}$ is the predicted count of the $i$-th class symbol. It is noteworthy that the feature maps of different branches contain different scale information and are highly complementary. Thus, we combine the complementary counting vectors and use the average operator to generate the final result $\mathcal{V}^f \in \mathbb{R}^{1 \times C}$, which is then fed into the decoder CCAD.

\subsection{Counting-Combined Attentional Decoder}\label{ccad}

% In the field of HMER, there are usually two types of decoder, sequential decoder and structural decoder. The latter one (e.g., DWAP-TD~\cite{DBLP:conf/icml/ZhangDYSW020}) decodes features in a structural way and needs to find parent node for each child node at every step, which inevitably brings a lot of extra time cost. Our counting-combined attentional decoder (CCAD) belongs to the sequential decoder in order to maintain a fast inference speed. 
% The comparison in parameters and inference speed between the two types of decoder is in section \ref{experiments}. 

\begin{figure*}[t]
\begin{center}
  \includegraphics[width=0.76\linewidth]{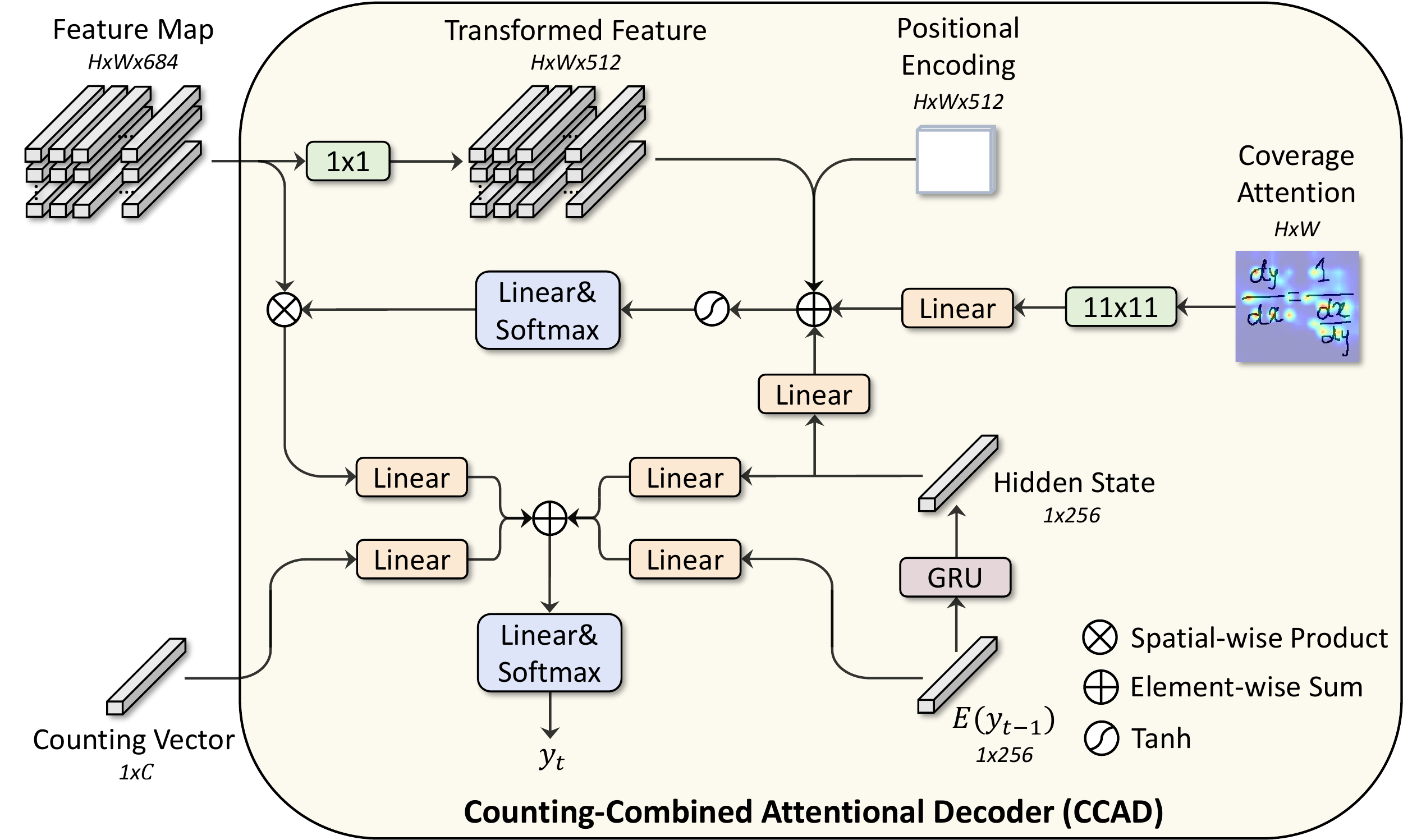}
\end{center}
\vspace{-1ex}
\caption{Structure of the proposed counting-combined attentional decoder (CCAD).}
\label{CCAD}
\end{figure*}

The structure of our counting-combined attentional decoder (CCAD) is shown in \fig{CCAD}. Given the 2D feature map $\mathcal{F} \in \mathbb{R}^{H \times W \times 684}$, we first use a $1 \times 1$ convolution to change the number of channel and get transformed feature $\mathcal{T} \in \mathbb{R}^{H \times W \times 512}$. Then, to enhance model's awareness of spatial position, we use a fixed absolute encoding $\mathcal{P} \in \mathbb{R}^{H \times W \times 512}$ to represent different spatial positions in $\mathcal{T}$. Specifically, we adopt the spatial positional encoding~\cite{DBLP:conf/icml/ParmarVUKSKT18}, which independently uses sine and cosine functions with different frequencies for both spatial coordinates.

During the decoding process, when decoding at step $t$, we pass the embedding of symbol $y_{t-1}$ into a GRU cell \cite{DBLP:conf/emnlp/ChoMGBBSB14} to get a hidden state $h_{t} \in \mathbb{R}^{1 \times 256}$. With the transformed feature $\mathcal{T}$ and the spatial encoding $\mathcal{P}$, we can then get the attention weights $\alpha_{t} \in \mathbb{R}^{H \times W}$ as follows:
% \vspace{+2ex}
\begin{equation}
% \vspace{+1ex}
\label{eq4}
\displaystyle{e_{t}=w^{T}\text{tanh}(\mathcal{T}+\mathcal{P}+W_{a}\mathcal{A}+W_{h}h_{t})+b, }
\vspace{-2.5ex}
\end{equation}

% \vspace{-3ex}
\begin{equation}
\label{eq5}
\displaystyle{\alpha_{t,ij}=\text{exp}(e_{t,ij})/\sum\limits_{p=1}^{\mathrm{H}}\sum\limits_{q=1}^{\mathrm{W}}e_{t,pq}, }
\vspace{+1ex}
\end{equation}
where $w$, $b$, $W_{a}$, $W_{h}$ are trainable weights and coverage attention $\mathcal{A}$ is the sum of all past attention weights.
% \vspace{-1ex}
% \begin{equation}
% \label{eq6}
% \displaystyle{\mathcal{A}=\sum\limits_{t^{'}=1}^{\mathrm{t-1}}\alpha_{t^{'}}}
% \end{equation}
% % \vspace{-2ex}

Applying spatial-wise product to the attention weights $\alpha_{t}$ and the feature map $\mathcal{F}$, we can get context vector $\mathcal{C} \in \mathbb{R}^{1 \times 256}$.
In most of the previous HMER methods, they predict $y_{t}$ only using the context vector $\mathcal{C}$, the hidden state $h_{t}$ and the embedding $E(y_{t-1})$. Actually, $\mathcal{C}$ just corresponds to a local region of the feature map $\mathcal{F}$. And we argue that $h_{t}$ and $E(y_{t-1})$ also lack global information. Considering that the counting vector $\mathcal{V}$ is calculated from a global counting perspective, which can serve as additional global information to make the prediction more accurate, we combine them together to predict $y_{t}$ as follows:
\begin{equation}
\vspace{+1.5ex}
\label{eq6}
\displaystyle{p(y_{t})=\text{softmax}(w_{o}^{T}(W_{c}\mathcal{C}+W_{v}\mathcal{V}+W_{t}h_{t}+W_{e}E)+b_{o}, }
\vspace{-4ex}
\end{equation}

\begin{equation}
\label{eq7}
\displaystyle{y_{t} \sim p(y_{t}), }
\vspace{+1ex}
\end{equation}
where $w_{o}$, $b_{o}$, $W_{c}$, $W_{v}$, $W_t$, $W_{e}$ are trainable weights. 

% and more specific impact of adding counting vector $\mathcal{V}$ is analysed in section \ref{ablation}.

\subsection{Loss Function}
The overall loss function consists of two parts and is defined as follows:
\begin{equation}
\label{eq8}
\displaystyle{\mathcal{L}=\mathcal{L}_{cls}+\mathcal{L}_{counting}, }
\end{equation}
where $\mathcal{L}_{cls}$ is a common-used cross entropy classification loss of the predicted probability $p(y_{t})$ with respect to its ground-truth. Denoting the counting ground truth of each symbol class as $\hat{\mathcal{V}}$, $\mathcal{L}_{counting}$ is a smooth $L1$~\cite{DBLP:journals/pami/RenHG017} regression loss defined as follows:
\vspace{+2ex}
\begin{equation}
\label{eq9}
\displaystyle{\mathcal{L}_{counting}=smooth_{L1}(\mathcal{V}, \hat{\mathcal{V}})}
\end{equation}

\section{Experiments}\label{experiments}

\subsection{Datasets}

\textbf{CROHME Dataset}~\cite{DBLP:conf/icfhr/MouchereVZG14} is the most widely-used public dataset in the field of HMER, which is from the competition on recognition of online handwritten mathematical expressions (CROHME). The CROHME training set contains 8836 handwritten mathematical expressions, and there are three testing sets: CROHME 2014, 2016, 2019 with 986, 1147, and 1199 handwritten mathematical expressions, respectively. The number of symbol classes $C$ is 111, including $``sos"$ and $``eos"$. In the CROHME dataset, each handwritten mathematical expression is stored in InkML format, which records the trajectory coordinates of handwritten strokes. We convert the handwritten stroke trajectory information in the InkML files into image format for training and testing.

\begin{figure*}[t]
\begin{center}
  \includegraphics[width=0.96\linewidth]{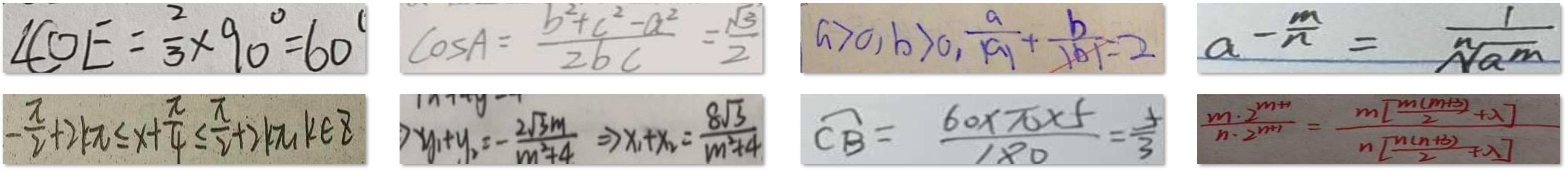}
\end{center}
\vspace{-2ex}
\caption{Some example images from the HME100K dataset.}
\label{hme}
% \vspace{-1ex}
\end{figure*}

% \vspace{+1ex}
\textbf{HME100K Dataset}~\cite{hme100k} is a real scene handwritten mathematical expression dataset, consisting of 74,502 images for training and 24,607 images for testing. The number of symbol classes $C$ is 249 including $``sos"$ and $``eos"$. These images are from tens of thousands of writers, mainly captured by cameras. Consequently, HME100K is more authentic and realistic with variations in color, blur, and complicated background. Some example images are shown in \fig{hme}.

\subsection{Implementation Details}

The proposed CAN is implemented in PyTorch. We use a single Nvidia Tesla V100 with 32GB RAM to train our model with batch size 8 and the Adadelta optimizer~\cite{DBLP:journals/corr/abs-1212-5701}. The learning rate starts from 0 and monotonously increases to 1 at the end of the first epoch and decays to 0 following the cosine schedules~\cite{DBLP:journals/corr/abs-1902-04103}. For the CROHME dataset, the total training epoch is set to 240, and we separately present the results with and without data augmentation. Compared with previous methods, we use different data augmentation (rotation, affine, perspective, erosion and dilation) to explore the ability of our method. For the HME100K dataset, the total training epoch is set to 30 without data augmentation. 

It is noteworthy that when counting the symbols in the handwritten mathematical expression, six classes of symbols are ignored by assigning their counting ground truth as zero because they are invisible: ``sos", ``eos", \text{``\^{}"}, \text{``\_{}"}, \text{``\{"}, \text{``\}"}. Counting these symbols will confuse the model and bring lower accuracy.

\subsection{Evaluation Metrics}

\textbf{Expression recognition}. Expression recognition rate (ExpRate), defined as the percentage of correctly recognized
expressions, is used to evaluate the performance of different methods on mathematical expression recognition. Moreover, $\leq1$ and $\leq2$ are also used, indicating that the expression recognition rate is tolerable at most one or two symbol-level errors.

% \vspace{+1ex}
\textbf{Symbol counting}. The mean absolute error (MAE) and the mean squared error (MSE) are the primary metrics in the object counting task. In our multi-class symbol counting task, we use MAE and MSE to evaluate the counting performance for each formula image, and then average the counting results of all formula images to get $MAE_{Ave}$ and $MSE_{Ave}$:
\begin{equation}
\label{eq10}
\displaystyle{MAE=\frac{1}{C} \sum\limits_{i=1}^{C}|\mathcal{V}_{i}-\hat{\mathcal{V}}_{i}|, \quad MSE=\sqrt{\frac{1}{C}\sum\limits_{i=1}^{C}|\mathcal{V}_{i}-\hat{\mathcal{V}}_{i}|^{2}},}
% \vspace{-1ex}
\end{equation}

\begin{equation}
\label{eq11}
\displaystyle{MAE_{Ave}=\frac{1}{N}\sum\limits_{i=1}^{N}MAE_{i}, \quad MSE_{Ave}=\frac{1}{N}\sum\limits_{i=1}^{N}MSE_{i},}
\vspace{+1ex}
\end{equation}
where $C$ denotes the number of symbol classes, $N$ is the number of images in the testing set, $\mathcal{V}_{i}$ and $\hat{\mathcal{V}}_{i}$ are the predicted count and its corresponding ground truth of a symbol class respectively.

% \underline{Underline} represents the second best result

\begin{table}[t]
\scriptsize
\caption{Results on the CROHME dataset. $^\star$ indicates using stoke trajectory coordinates annotations. $^\dagger$ indicates our reproduced result. CAN-DWAP and CAN-ABM represent using DWAP and ABM as the baseline respectively. Note that we use different data augmentation than previous methods. Our intention is to show that even with data augmentation, our counting module can still stably improve existing HMER methods’ performance.}
% \vspace{+1ex}
\centering
\begin{tabular}{|m{.24\columnwidth}|m{.10\columnwidth}<{\centering}|m{.06\columnwidth}<{\centering}|m{.06\columnwidth}<{\centering}|m{.10\columnwidth}<{\centering}|m{.06\columnwidth}<{\centering}|m{.06\columnwidth}<{\centering}|m{.10\columnwidth}<{\centering}|m{.06\columnwidth}<{\centering}|m{.06\columnwidth}<{\centering}|}
\hline
\multirow{2}*{Method}& \multicolumn{3}{c|}{CROHME 2014}&  \multicolumn{3}{c|}{CROHME 2016}& \multicolumn{3}{c|}{CROHME 2019}\\
\cline{2-10}
 & ExpRate$\uparrow$ & $\leq1\uparrow$ & $\leq2\uparrow$ & ExpRate$\uparrow$ & $\leq1\uparrow$ & $\leq2\uparrow$ & ExpRate$\uparrow$ & $\leq1\uparrow$ & $\leq2\uparrow$\\
\hline\hline
\rowcolor{gray!20}
\multicolumn{10}{|l|}{Without data augmentation}\\ 
\hline
UPV~\cite{DBLP:conf/icfhr/MouchereVZG14} & 37.22 & 44.22 & 47.26 & - & - & - & - & - & -\\
TOKYO~\cite{DBLP:conf/icfhr/MouchereVZG16} & - & - & - & 43.94 & 50.91 & 53.70 & - & - & -\\
PAL~\cite{DBLP:conf/pkdd/WuYZZL18} & 39.66 & 56.80 & 65.11 & - & - & - & - & - & -\\
WAP~\cite{DBLP:journals/pr/ZhangDZLHHWD17} & 46.55 & 61.16 & 65.21 & 44.55 & 57.10 & 61.55 & - & - & -\\
% PGS~\cite{DBLP:journals/prl/LeIN19} & 48.78 & 66.13 & 73.94 & 36.27 & - & - & - & - & -\\
PAL-v2~\cite{DBLP:journals/ijcv/WuYZZL20} & 48.88 & 64.50 & 69.78 & 49.61 & 64.08 & 70.27 & - & - & -\\
TAP~\cite{DBLP:journals/tmm/ZhangDD19}$^\star$ & 48.47 & 63.28 & 67.34 & 44.81 & 59.72 & 62.77 & - & - & -\\
DLA~\cite{DBLP:conf/cvpr/Le20} & 49.85 & - & - & 47.34 & - & - & - & - & -\\
DWAP~\cite{DBLP:conf/icpr/ZhangDD18} & 50.10 & - & - & 47.50 & - & - & - & - & -\\
DWAP-TD~\cite{DBLP:conf/icml/ZhangDYSW020} & 49.10 & 64.20 & 67.80 & 48.50 & 62.30 & 65.30 & 51.40 & 66.10 & 69.10\\
DWAP-MSA~\cite{DBLP:conf/icpr/ZhangDD18} & 52.80 & 68.10 & 72.00 & 50.10 & 63.80 & 67.40 & 47.70 & 59.50 & 63.30\\
WS-WAP~\cite{DBLP:conf/icfhr/TruongNPN20} & 53.65 & - & - & 51.96 & 64.34 & 70.10 & - & - & -\\
MAN~\cite{DBLP:conf/icdar/WangDZW19}$^\star$ & 54.05 & 68.76 & 72.21 & 50.56 & 64.78 & 67.13 & - & - & -\\
BTTR~\cite{DBLP:conf/icdar/ZhaoGYPDZ21} & 53.96 & 66.02 & 70.28 & 52.31 & 63.90 & 68.61 & 52.96 & 65.97 & 69.14\\
ABM~\cite{abm} & 56.85 & 73.73 & 81.24 & 52.92 & 69.66 & 78.73 & 53.96 & 71.06 & 78.65\\
% DWAP-TD-v2 & 53.62 & - & - & \underline{55.18} & - & - & \textbf{58.72} & - & -\\
% SAN & 56.20 & 72.60 & 79.20 & 53.60 & 69.60 & 76.80 & - & - & -\\
\hline
DWAP (baseline)$^\dagger$ & 51.48 & 67.01 & 73.30 & 50.65 & 63.30 & 70.88 & 50.04 & 65.39 & 69.39\\
CAN-DWAP (ours) & \textbf{57.00} & 74.21 & 80.61 & \textbf{56.06} & 71.49 & 79.51 & \textbf{54.88} & 71.98 & 79.40\\
\hline
ABM (baseline)$^\dagger$ & 56.04 & 73.10 & 79.90 & 53.36 & 70.01 & 78.12 & 53.71 & 71.23 & 78.23\\
CAN-ABM (ours) & \textbf{57.26} & 74.52 & 82.03 & \textbf{56.15} & 72.71 & 80.30 & \textbf{55.96} & 72.73 & 80.57\\
\hline
\hline\hline
\rowcolor{gray!20}
\multicolumn{10}{|l|}{With data augmentation}\\ 
\hline
Li \emph{et al.}~\cite{DBLP:conf/icfhr/LiJLZ20} & 56.59 & 69.07 & 75.25 & 54.58 & 69.31 & 73.76 & - & - & -\\
Ding \emph{et al.}~\cite{DBLP:conf/icdar/Ding0H21} & 58.72 & - & - & 57.72 & 70.01 & 76.37 & 61.38 & 75.15 & 80.23\\
\hline
DWAP (baseline)$^\dagger$ & 57.97 & 73.81 & 79.19 & 55.97 & 71.40 & 79.86 & 56.05 & 72.23 & 79.15\\
CAN-DWAP (ours) & \textbf{65.58} & 77.36 & 83.35 & \textbf{62.51} & 74.63 & 82.48 & \textbf{63.22} & 78.07 & 82.49\\
\hline
ABM (baseline)$^\dagger$ & 63.76 & 76.35 & 83.05 & 60.86 & 73.93 & 81.17 & 62.22 & 77.23 & 81.90\\
CAN-ABM (ours) & \textbf{65.89} & 77.97 & 84.16 & \textbf{63.12} & 75.94 & 82.74 & \textbf{64.47} & 78.73 & 82.99\\
\hline
\end{tabular}
\label{sota}
\end{table}

\begin{figure*}[!ht]
\begin{center}
  \includegraphics[width=0.86\linewidth]{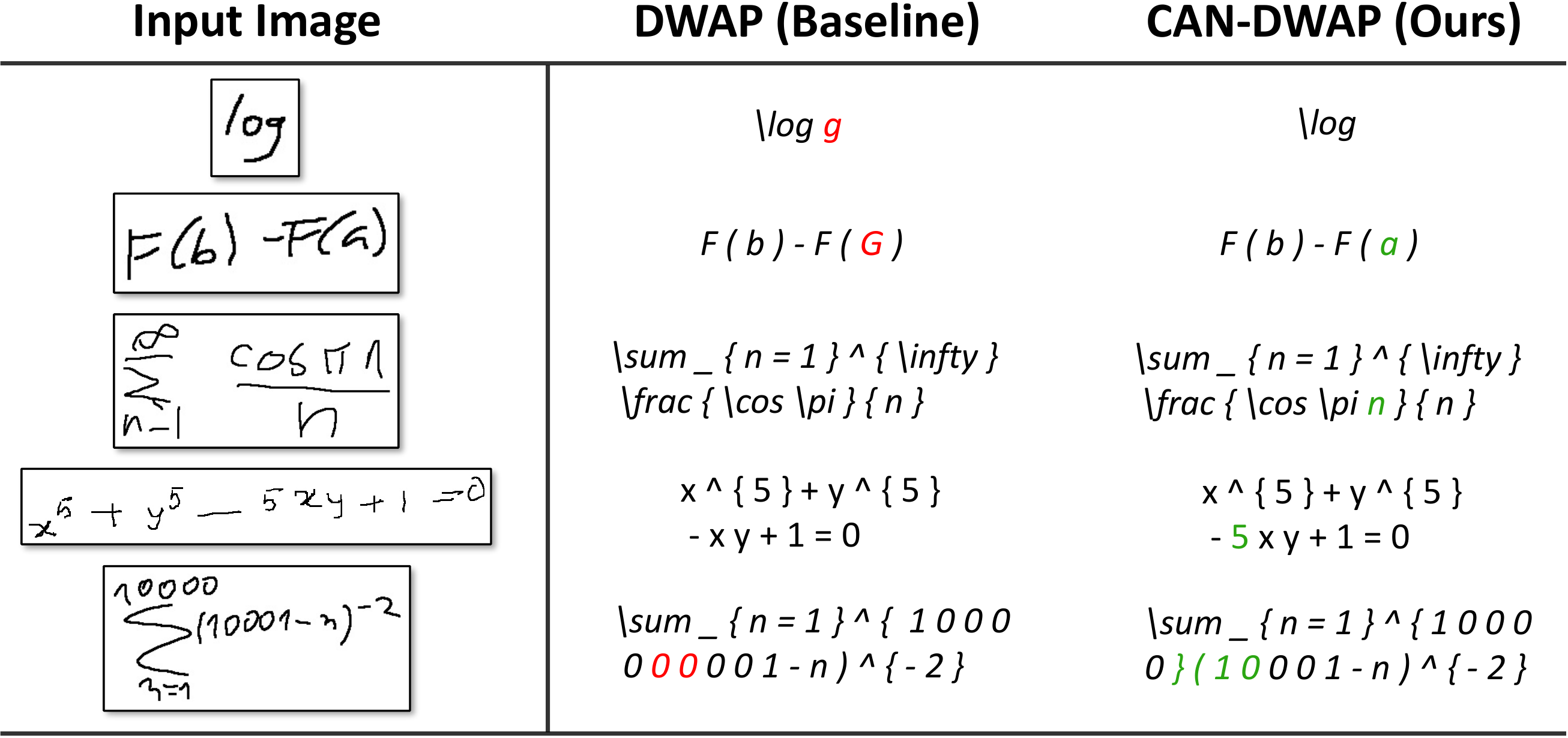}
\end{center}
\vspace{-1.5ex}
\caption{Some recognition cases of DWAP and CAN-DWAP.}
\label{example}
% \vspace{-2ex}
\end{figure*}

\subsection{Comparison with State-of-the-Art}
% \vspace{-2ex}
To demonstrate the superiority of our method, we compare it with previous state-of-the-art (SOTA) methods. \tab{sota} shows the expression recognition rate (ExpRate) on the CROHME dataset. Most of the previous methods do not use data augmentation, so we mainly focus on the results produced without data augmentation. 

As shown in \tab{sota}, with adopting DWAP~\cite{DBLP:conf/icpr/ZhangDD18} as the baseline, CAN-DWAP achieves SOTA results on CROHME 2014, CROHME 2016, CROHME 2019 and outperforms the latest SOTA method ABM~\cite{abm} on CROHME 2016 by a significant margin of 3.14\%. \fig{example} shows some qualitative recognition results of DWAP and CAN-DWAP. We can observe that our method is less likely to miss symbols or predict redundant symbols. 
% Compared with the baseline DWAP~\cite{DBLP:conf/icpr/ZhangDD18}, the improvement of ExpRate is significant: 5.52\% on CROHME 2014 (from 51.48\% to 57.00\%), 5.41\% on CROHME 2016 (from 50.65\% to 56.06\%) and 4.84\% on CROHME 2019 (from 50.04\% to 54.88\%).

% Moreover, CAN-DWAP largely outperforms WS-WAP~\cite{DBLP:conf/icfhr/TruongNPN20}, which predicts 0/1 to represent if a class of symbol exists in the image or not. This demonstrates that compared with symbol classification, symbol counting is more beneficial to HMER. We think the reason behind is that symbol counting is a more high-level task than symbol classification, since counting a class of symbols needs to capture all symbols of this class. While for classification, just capturing one symbol of this class is enough for the model to make prediction.

To further verify the effectiveness of our method, we reproduce the latest SOTA method ABM~\cite{abm} and adopt it as our baseline to construct CAN-ABM. As shown in \tab{sota}, CAN-ABM outperforms its baseline and achieves new SOTA results. This indicates that our method can be generalized to various existing encoder-decoder models for HMER and boost their performance.

\subsection{Results on the HME100K Dataset}\label{hme100k}

Although the CROHME dataset has been widely used and has great influence in the field of HMER, its small size limits the performance of different methods. Hence, we further evaluate our method on the HME100K dataset, which is nearly ten times larger than the CROHME dataset and has more variations in color, blur, and background. The quantitative results are listed in \tab{hme_result}, CAN-DWAP and CAN-ABM largely outperform their baseline DWAP~\cite{DBLP:conf/icpr/ZhangDD18} and ABM~\cite{abm} respectively.

% \vspace{-2ex}
\subsection{Inference Speed}\label{speed}
% \vspace{-1ex}
To explore the efficiency of our proposed method, we evaluate its speed on the HME100K dataset with a single Nvidia Tesla V100. As shown in \tab{fps}, compared with the baseline model, the extra parameters and FLOPs are mainly brought by the counting module’s two convolution layers with kernels of sizes $3 \times 3$ and $5 \times 5$. As to the inference speed, the extra time cost brought by the counting module is marginal.

% \vspace{-1ex}
\subsection{Ablation Study}\label{ablation}
% \vspace{-1ex}
\textbf{Component Analysis.}
In our method, symbol counting serves as an auxiliary task and influences the feature learning together with the primary task HMER through joint optimization. Meanwhile, adding counting vector during the decoding process also has an impact on the performance. So, to verify the effectiveness of the three components: positional encoding, joint optimization, and counting vector, we conduct experiments and the results are listed in \tab{components}. We can observe that both joint optimization and counting vector can boost the performance to a certain degree, and adding positional encoding can also slightly improve the recognition accuracy. 

\begin{table}[t]
% \vspace{+1ex}
\scriptsize
\caption{Results on the HME100K dataset. $^\dagger$ indicates our reproduced result. CAN-DWAP and CAN-ABM represent using DWAP and ABM as the baseline respectively.}
% \vspace{+1ex}
\centering
\begin{tabular}{|m{.25\columnwidth}|m{.12\columnwidth}<{\centering}|m{.12\columnwidth}<{\centering}|m{.12\columnwidth}<{\centering}|}
\hline
\multirow{2}*{Method}& \multicolumn{3}{c|}{HME100K}\\
\cline{2-4}
 & ExpRate$\uparrow$ & $\leq1\uparrow$ & $\leq2\uparrow$\\
\hline\hline
DWAP-TD~\cite{DBLP:conf/icml/ZhangDYSW020}$^\dagger$ & 62.60 & 79.05 & 85.67\\
\hline
DWAP~\cite{DBLP:conf/icpr/ZhangDD18} (baseline)$^\dagger$ & 61.85 & 70.63 & 77.14\\
CAN-DWAP (ours) & 67.31 & 82.93 & 89.17\\
\hline
ABM~\cite{abm} (baseline)$^\dagger$ & 65.93 & 81.16 & 87.86\\
CAN-ABM (ours) & \textbf{68.09} & \textbf{83.22} & \textbf{89.91}\\
\hline
\end{tabular}
% \vspace{-3ex}
\label{hme_result}
\vspace{+1ex}
\end{table}

\begin{table}[t]
% \vspace{+1ex}
\scriptsize
\caption{Comparison on parameters, FLOPs, and FPS.}
% \vspace{+1ex}
\centering
\begin{tabular}{|m{.24\columnwidth}|m{.12\columnwidth}<{\centering}|m{.18\columnwidth}<{\centering}|m{.12\columnwidth}<{\centering}|m{.12\columnwidth}<{\centering}|}
\hline
\multirow{2}*{Method}& \multicolumn{4}{c|}{HME100K}\\
\cline{2-5}
 & \#Params & Input size & FLOPs & FPS\\
\hline\hline
DWAP~\cite{DBLP:conf/icpr/ZhangDD18} (baseline) & 4.7M & (1,1,120,800) & 9.7G & 23.3\\
CAN-DWAP (ours) & 17.0M & (1,1,120,800) & 14.7G & 21.7\\
\hline
\end{tabular}
% \vspace{-3ex}
\label{fps}
% \vspace{+2ex}
\end{table}

% \vspace{+1ex}
\textbf{Impact of Convolution Kernel in Counting Module.}
In our counting module MSCM, we adopt a multi-scale strategy by using convolution layer with different sizes of kernels ($3 \times 3$ and $5 \times 5$). To explore the impact of different convolution kernels, we conduct experiments on CROHME 2014 with using different sizes of convolution kernels. As shown in \tab{filters}, using $3 \times 3$ and $5 \times 5$ convolution kernels together achieves the best results (57.00\% ExpRate, 0.033 $MAE_{Ave}$ and 0.037 $MSE_{Ave}$). Using either $3 \times 3$ or $5 \times 5$ convolution kernel will get lower counting accuracy and lower ExpRate. We think this phenomenon indicates that multi-scale information obtained with different kinds of convolution kernels can help the counting module better tackle the size variations.     

\begin{table}[t]
\scriptsize
\caption{Ablation study of different components.}
% \vspace{-1ex}
\centering
\begin{tabular}{|m{.24\columnwidth}|m{.10\columnwidth}<{\centering}|m{.06\columnwidth}<{\centering}|m{.06\columnwidth}<{\centering}|m{.10\columnwidth}<{\centering}|m{.06\columnwidth}<{\centering}|m{.06\columnwidth}<{\centering}|m{.10\columnwidth}<{\centering}|m{.06\columnwidth}<{\centering}|m{.06\columnwidth}<{\centering}|}
\hline
\multirow{2}*{Method}& \multicolumn{3}{c|}{CROHME 2014}&  \multicolumn{3}{c|}{CROHME 2016}& \multicolumn{3}{c|}{CROHME 2019}\\
\cline{2-10}
& ExpRate$\uparrow$ & $\leq1\uparrow$ & $\leq2\uparrow$ & ExpRate$\uparrow$ & $\leq1\uparrow$ & $\leq2\uparrow$ & ExpRate$\uparrow$ & $\leq1\uparrow$ & $\leq2\uparrow$\\
\hline\hline
DWAP~\cite{DBLP:conf/icpr/ZhangDD18} (baseline) & 51.48 & 67.01 & 73.30 & 50.65 & 63.30 & 70.88 & 50.04 & 65.39 & 69.39\\
+ Positional encoding & 51.88 & 68.12 & 74.21 & 51.00 & 64.06 & 71.37 & 50.96 & 66.14 & 70.48\\
+ Joint optimization & 55.23 & 72.18 & 78.17 & 54.11 & 68.00 & 76.37 & 53.13 & 69.89 & 76.00\\
+ Counting vector & \textbf{57.00} & \textbf{74.21} & \textbf{80.61} & \textbf{56.06} & \textbf{71.49} & \textbf{79.51} & \textbf{54.88} & \textbf{71.98} & \textbf{79.40}\\
\hline
\end{tabular}
\label{components}
% \vspace{+2ex}
\end{table}

% \vspace{-3ex}
\begin{table}[!t]
\scriptsize
\caption{Ablation study of different convolution kernels in counting module.}
% \vspace{-1ex}
\centering
\begin{tabular}{|m{.30\columnwidth}|m{.12\columnwidth}<{\centering}|m{.12\columnwidth}<{\centering}|m{.12\columnwidth}<{\centering}|m{.12\columnwidth}<{\centering}|m{.12\columnwidth}<{\centering}|}
\hline
\multirow{2}*{Method}& \multicolumn{5}{c|}{CROHME 2014}\\
\cline{2-6}
 & ExpRate$\uparrow$ & $\leq1\uparrow$ & $\leq2\uparrow$ & $MAE_{Ave}\downarrow$ & $MSE_{Ave}\downarrow$\\
\hline\hline
CAN-DWAP ($3 \times 3$) & 54.92 & 71.26 & 78.07 & 0.048 & 0.046\\
CAN-DWAP ($5 \times 5$) & 55.53 & 71.88 & 78.58 & 0.044 & 0.043\\
CAN-DWAP ($3 \times 3$ \& $5 \times 5$) & \textbf{57.00} & \textbf{74.21} & \textbf{80.61} & \textbf{0.033} & \textbf{0.037}\\
\hline
\end{tabular}
\label{filters}
% \vspace{+1ex}
\end{table}

\begin{table}[!t]
\vspace{+1.5ex}
\scriptsize
\caption{Ablation study of different counting vectors. $^\ast$ indicates adding random disturbance to counting vector. The latter counting GT with $^\ast$ is added with more disturbances than the former one.}
% \vspace{-1ex}
\vspace{+0.5ex}
\centering
\begin{tabular}{|m{.30\columnwidth}|m{.12\columnwidth}<{\centering}|m{.12\columnwidth}<{\centering}|m{.12\columnwidth}<{\centering}|m{.12\columnwidth}<{\centering}|m{.12\columnwidth}<{\centering}|}
\hline
\multirow{2}*{Method}& \multicolumn{5}{c|}{CROHME 2014}\\
\cline{2-6}
 & ExpRate$\uparrow$ & $\leq1\uparrow$ & $\leq2\uparrow$ & $MAE_{Ave}\downarrow$ & $MSE_{Ave}\downarrow$\\
\hline\hline
CAN-DWAP & 57.00 & 74.21 & 80.61 & 0.033 & 0.037\\
CAN-DWAP (counting GT)$^\ast$ & 58.28 & 74.92 & 81.02 & 0.027 & 0.025\\
CAN-DWAP (counting GT)$^\ast$ & 60.10 & 76.04 & 81.73 & 0.019 & 0.016\\
CAN-DWAP (counting GT) & 62.44 & 76.14 & 82.23 & 0.000 & 0.000\\
\hline
\end{tabular}
% \vspace{-3ex}
\label{vector}
% \vspace{-2ex}
\end{table}

% \vspace{+1ex}
\textbf{Impact of Counting Vector on HMER.} 
% As described in \eq{eq6} and validated above, when predicting at each step, adding counting vector can improve the ExpRate. And the more accurate the counting vector is (measured with $MAE_{Ave}$ and $MSE_{Ave}$), the higher the ExpRate is. 
To explore the impact of counting vector, we use the ground truth of counting vector and add different random disturbances to it (e.g., randomly add or subtract 1) so that we can get several counting vectors with different $MAE_{Ave}$ and $MSE_{Ave}$. By providing these counting vectors to the decoder during training and testing, we conduct several experiments and the results are shown in \tab{vector}. When using the ground truth of counting vector, the ExpRate on CROHME 2014 reaches 62.44\%. As more disturbances are added, the counting vector becomes more inaccurate, and the ExpRate drops consequently.   

\begin{table}[t]
\scriptsize
\caption{Ablation study of HMER's impact on symbol counting.}
% \vspace{+1ex}
\centering
\begin{tabular}{|m{.24\columnwidth}|m{.14\columnwidth}<{\centering}|m{.14\columnwidth}<{\centering}|}
\hline
\multirow{2}*{Method}& \multicolumn{2}{c|}{CROHME 2014}\\
\cline{2-3}
 & $MAE_{Ave}\downarrow$ & $MSE_{Ave}\downarrow$\\
\hline\hline
Counting w/o HMER & 0.048 & 0.044\\
Counting w HMER & \textbf{0.033} & \textbf{0.037}\\
\hline
\end{tabular}
\label{impact}
% \vspace{+1ex}
\end{table}

\begin{figure*}[t]
\vspace{+1ex}
\begin{center}
  \includegraphics[width=0.76\linewidth]{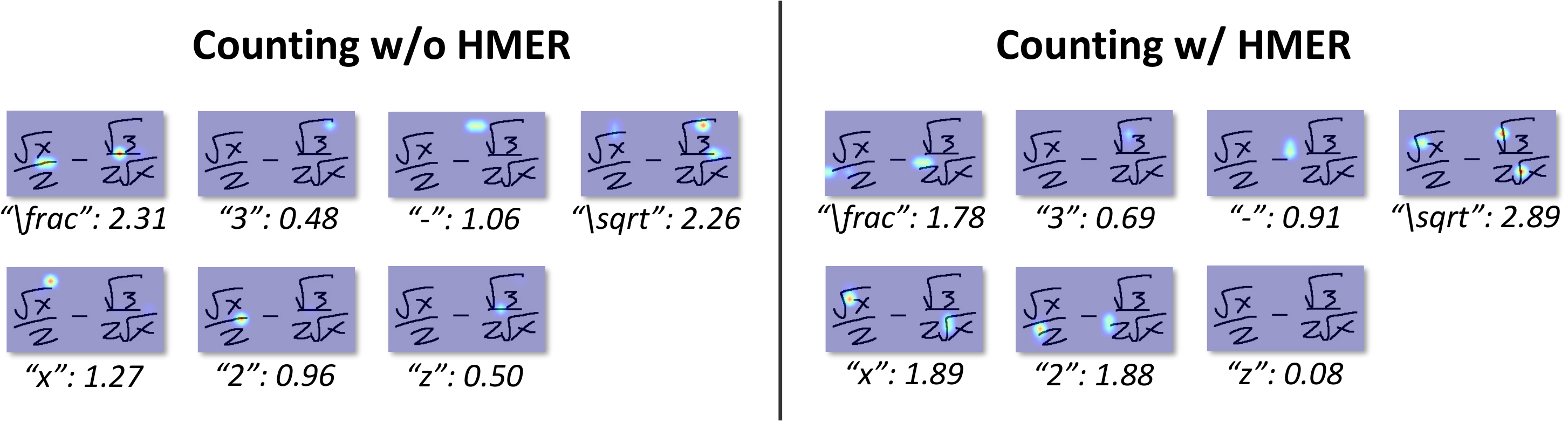}
\end{center}
\vspace{-3ex}
\caption{Counting map generated with and without HMER task.}
\label{counting}
\vspace{+3ex}
\end{figure*}

\begin{figure*}[!t]
\begin{center}
  \includegraphics[width=0.94\linewidth]{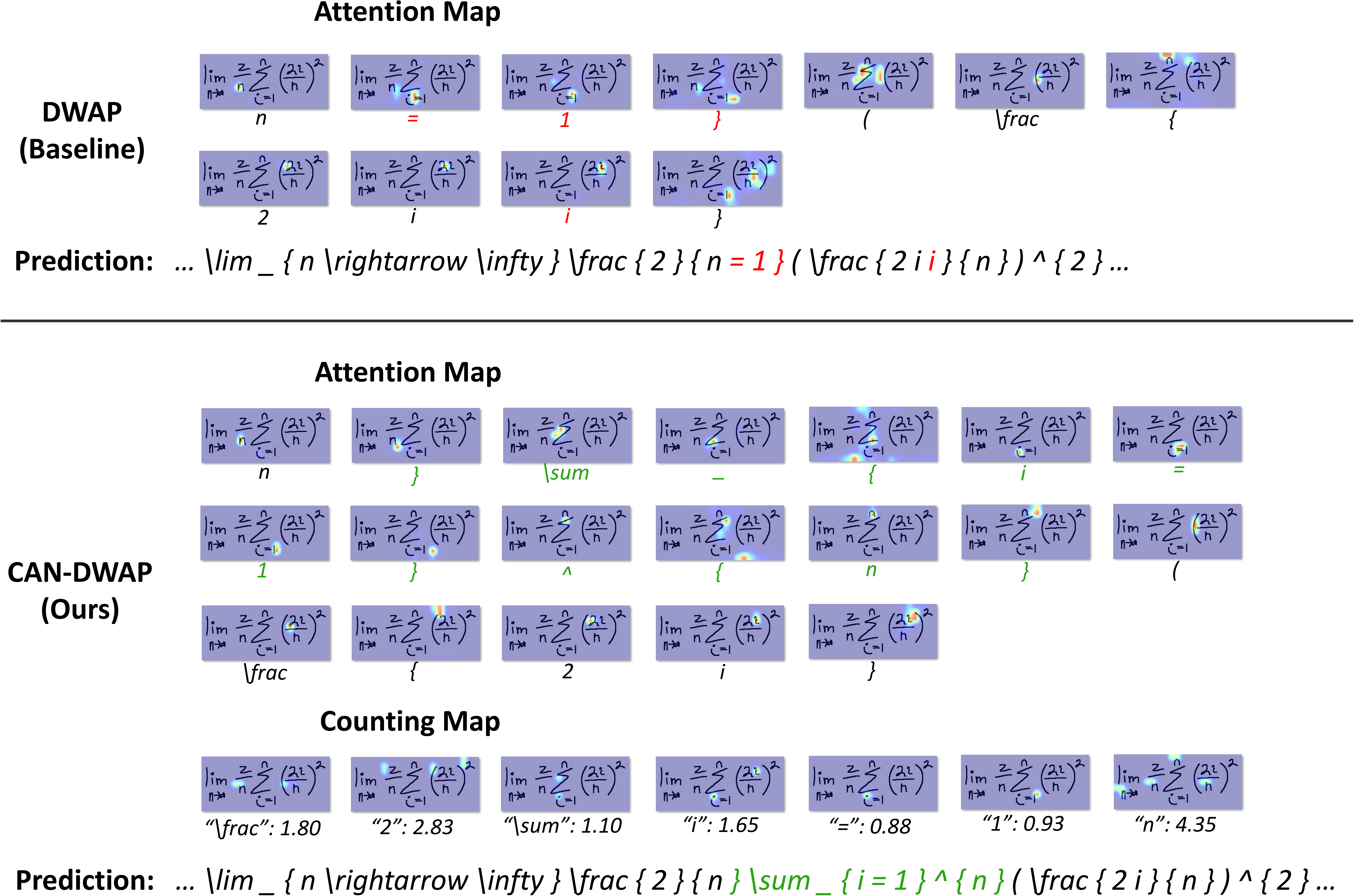}
\end{center}
\vspace{-1.5ex}
\caption{Counting map and attention map of DWAP and CAN-DWAP.}
\label{map}
\end{figure*}

% \vspace{+1ex}
\textbf{Impact of HMER on Symbol Counting.}
Through joint optimization, symbol counting can promote the performance of HMER. To find out whether HMER can also promote the performance of symbol counting, we train CAN only with the symbol counting task and compare it with CAN trained with two tasks. As shown in~\tab{impact}, HMER can boost the performance of symbol counting with improving $MAE_{Ave}$ by 31.25\% and $MSE_{Ave}$ by 15.91\%. Some visual results are shown in \fig{counting}. We can observe that when training only with the symbol counting task, some symbols are wrongly located (e.g., $``-"$) or partially counted (e.g., $``2"$). Counting with the HMER task can alleviate this problem by providing context-aware information, which is gained through the context-aware decoding process in the decoder CCAD.

\subsection{Case Study with Maps}\label{case_study}

In this part, we choose a typical example to visualize its counting map from the counting module and its attention map from the decoder. As illustrated in \fig{map}, after predicting the symbol $``n"$, DWAP misses the symbol $``\sum”$ and the symbol $``i"$ and directly predicts the symbol $``="$. The missing symbol $``\sum”$ is noticed later by the model when predicting the symbol $``("$ but the mistake has already happened in this sequential decoding process. A redundant symbol $``i"$ is also wrongly predicted, and the attention map shows that this mistake is due to the model's repeated attention on the symbol $``i"$.

In contrast, our method CAN-DWAP predicts the formula correctly. From the counting map, we can see that almost all symbols are accurately located (note that we do not use symbol-level position annotations). And the predicted count of each symbol class, which is calculated by summing each counting map, is very close to its ground truth. These phenomenons demonstrate that by counting each symbol class, the model becomes more aware of each symbol, especially their positions. As a result, the model has more accurate attention results (seen from the attention map) during the decoding process and is less likely to miss or predict redundant symbols.

% \vspace{-1ex}

\subsection{Limitation}\label{limitation}

\begin{figure*}[t]
\begin{center}
  \includegraphics[width=0.88\linewidth]{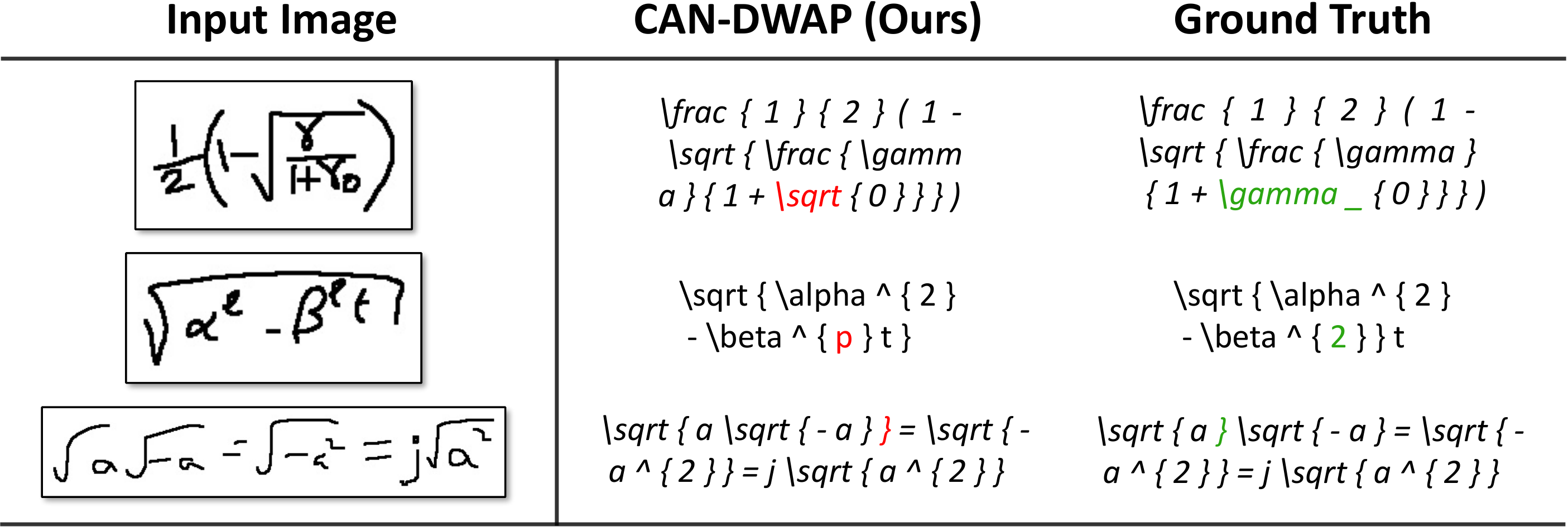}
\end{center}
\vspace{-2ex}
\caption{Some failure cases of our CAN-DWAP.}
\label{example_f}
% \vspace{-1ex}
\end{figure*}

Despite the significant performance improvement brought by symbol counting, the variations in writing styles still cause some recognition problems and cannot be solved very well with symbol counting, as shown in \fig{example_f}. Moreover, since we do not explicitly model structure grammar, our method may make some mistakes when extreme fine structure perception ability is needed.

\section{Conclusion}\label{conclusion}

In this paper, we design a counting module MSCM, which can perform symbol counting just relying on the original HMER annotations (LaTeX sequences). By plugging this counting module into an attention-based encoder-decoder network, we propose an unconventional end-to-end trainable network for HMER named CAN, which jointly optimizes HMER and symbol counting. Experiments on the benchmark datasets for HMER validate three main conclusions. 1) Symbol counting can consistently improve the performance of the encoder-decoder models for HMER. 2) Both joint optimization and counting results contribute to this improvement. 3) HMER can also increase the accuracy of symbol counting through joint optimization. 

% We hope this paper can provide a new perspective to promote the performance of HMER. In the future, we are interested in adopting symbol counting in other image-to-sequence tasks such as chemical formula recognition.  

\section*{Acknowledgements}

This work was done when Bohan Li was an intern at Tomorrow Advancing Life, and was supported in part by the National Natural Science Foundation of China 61733007 and the National Key R\&D Program of China under Grant No. 2020AAA0104500. 

\clearpage
\bibliographystyle{splncs04}
\bibliography{egbib}
\end{document}